# Morphological Analysis of the Bishnupriya Manipuri Language using Finite State Transducers

Nayan Jyoti Kalita[1], Navanath Saharia[1], and Smriti Kumar Sinha[1]

Tezpur University, India, PIN-704028

**Abstract.** In this work we present a morphological analysis of Bishnupriya Manipuri language, an Indo-Aryan language spoken in the north eastern India. As of now, there is no computational work available for the language. Finite state morphology is one of the successful approaches applied in a wide variety of languages over the year. Therefore we adapted the finite state approach to analyse morphology of the Bishnupriya Manipuri language.

## 1 Introduction

Morphology involves the study of inner structures of words and their different forms and is an essential step to any natural language processing task. The input to a morphological analyser is a word and it gives the lexical form of the input words, consisting of the root of the word and other morphological information the word has as output. There are two challenges in the morphological analysis; the morphotactics and the orthography. The morphotactics show how the morphemes combine together to form a word. The orthography is the model of how spelling changes when the morphemes combine together. One of the most efficient approaches to morphological analysis is the finite state transducer (FST) approach which determines whether an input string of morphemes makes up a word or not. [1] successfully implemented the finite state approach to morphology. He combined both the morphotactics and the orthography into a single lexical transducer. There are number of tools available for the construction of FST based morphological analyser among which XFST[1], SFST[2] and OFST[3] are popular among researchers. Our experiment in this work is based on XFST tool. In this work, we report the development of a morphological analyser using finite state transducers on Bishnupriya Manipuri. The Bishnupriya Manipuri language is an Indo-Aryan language spoken in the boarder of Tibeto-Burman language

---

[1] Xerox Finite State Transducer;
http://www.fsmbook.com
[2] Stuttgart Finite State Transducer;
http://www.cis.uni-muenchen.de/∼schmid/tools/SFST
[3] OpenFST;
http://www.openfst.org/twiki/bin/view/FST/webhome

family. We feel, being a boarder language and a resource scare language this work is significant.

The organization of the paper is as follows. Section 2 gives an outline on the Bishnupriya Manipuri language. Section 3 describes the related work on the morphological analysis. Section 4 describes the corpus development. Section 5 gives the architecture of the experiment. Section 6 gives the results obtained from our experiment. Section 7 concludes the report.

## 2 The Bishnupriya Manipuri Language

The Bishnupriya Manipuri language is of Indo-Aryan origin and is a kin to Assamese, Bengali and Oriya and neighbour of the Meitei language[4]. Being a boarder language of another language family, the Bishnupriya Manipuri language has around 4,000 borrowed root words originated from Meitei. Though the speaker of the language was originally confined only to the surroundings of the lake Loktak in Manipur of India, it is practically dead in its place of origin [2] . However, the language is retained by its speakers in diaspora mostly in Assam, Tripura, a few in Manipur of India and Bangladesh According to Ethnologue[5], total number of speakers of this language is approx. 117500, where 77,500 speakers are from India (2001 census). There are two dialects in the Bishnupriya Manipuri language, namely, the *Madai Gang* or the dialect of the village of the queen and the *Rajar Gang* or the dialect of the village of the king. Being an agglutinative language, Bishnupriya Manipuri has the capability of generating hundreds of words from a single noun and verb root. For example, the root word মানু (**manu** : *man*) may form different inflected words.

1. মানুয়ে (**manuje**) → মানু (**manu**) + য়ে (**je**)
   → Noun + NCM
2. মানুহাবি (**manuhabi**) → মানু (**manu**) + হাবি (**habi**)
   → Noun + Pl
3. মানুরে (**manuɹe**) → মানু (**manu**) + রে (**ɹe**)
   → Noun + ACM
4. মানুগয়ে (**manugɔje**) → মানু (**manu**) + গ (**gɔ**) + য়ে (**je**)
   → Noun + DM + NCM
5. মানুহাবিয়েহে (**manuhabijehe**) → মানু (**manu**) + হাবি (**habi**) + য়ে (**je**) + হে (**he**)
   → Noun + Pl + NCM + EM

A noun root can take case marker (locative, genitive, ablative, nominative, accusative, vocative, instrumental, dative), emphatic marker, number (singular

---
[4] A language of the Kuki-Chin branch of the Tibeto-Burman language family
[5] http://www.ethnologue.com/language/bpy

and plural) marker and definitive marker in the form of sequence of suffix. Likewise, a verb root may take tense marker (simple present, simple past, simple future), aspect marker, person ($1^{st}$, $2^{nd}$, $3^{rd}$) marker and mood (indicative, imperative, subjunctive) marker as suffix. Table 1 shows some examples of inflections in the Bishnupriya Manipuri language. Table 2 shows the verbal inflection of root কর (kɔɹ : *to do*)

During the study we found a peculiar characteristic of this language. The Bishnupriya Manipuri language is an Indo-Aryan language. Since it was developed in the context of Tibeto-Burman languages, it has borrowed a number of Tibeto-Burman noun and verb roots, mainly from Meitei. Interestingly, the Bishnupriya Manipuri suffixes cannot be added directly with the Meitei verb roots. An Indo-Aryan verb root is to be added first with the Meitei verb root forming this basis for affixation. For example-

হংকরানি (hɔŋkɔɹani : *to make*)→ হং (*Meitei root*)+ কর (Sanskrit/Indo-Aryan root)+ আনি (*suffix*)

| Words (IPA) | Output |
|---|---|
| দাদাগাছি (dadagaʃi) | দাদা_Noun+Pl |
| মাছগি (maʃgi) | মাছ_Noun+PDM |
| গুরুমাহেই (guɹumahei) | গুরু_Noun+Pl |
| মাছগুলি (maʃguli) | মাছ_Noun+Pl |
| মানুহাবি (manuhabi) | মানু_Noun+Pl |
| ঘরে (gʰɔɹe) | ঘরে_Noun+Sg+LCM |
| পাছিলে (paʃile) | পা_Verb+PsPSg |
| পাছিলায় (paʃilai) | পা_Verb+PsPPl |

**Table 1.** Examples of inflections in Bishnupriya Manipuri language. PDM→Plural Definitive Marker; Pl→Plural Marker; LCM→Locative Case Marker; PsPSg→Past-Perfect-Singular; PsPPl→Past-Perfect-Plural Marker

|  | $1^{st}$ person | $2^{nd}$ person | $3^{rd}$ person |
|---|---|---|---|
| Present | করুরি (kɔɹuɹi) | করর (kɔɹɔɹ) | করের (kɔɹeɹ) |
| Past | করলু (kɔɹlu) | করলে (kɔɹle) | করল (kɔɹlɔ) |
| Future | করতৌ (kɔɹtou) | করতেই (kɔɹtei) | করতই (kɔɹtɔi) |

**Table 2.** Inflectional forms of the verb কর (kɔɹ : *to do*) with tense and person.

## 3 Related Works

Number of researches has been done for developing morphological analyser in various languages based on finite state transducers. [3] showed how the morpho-tactics and the variation rules of Arabic have been described using only finite state operation and implemented a significant morphological analyser using this approach. [4] designed a morphological analyser and generator for Aymara language using Xerox finite-state machine. [5] described some of the challenges encountered in a computational morphological analysis of Persian and developed a morphological analyser for Persian using Xerox finite-state tools.

In Indian language context, [6] developed a morphological analyser for Hindi. They used SFST tool for generating the finite states. A morphological generator was developed for Kannada language using FST [7]. [8] described a finite state morphological analyser for the nouns in Oriya. He specified the co-occurrence restrictions of the morphemes in a nominal form and used the FSA to determine whether an input string of morphemes makes up an Oriya noun or not. [9], reported a paradigm-based finite state morphological analyser for Marathi language. Among languages spoken in north eastern India, we do not find any work on morphological analysis using FST.

## 4  Bishnupriya Manipuri Corpus

For the morphological analysis of the Bishnupriya Manipuri language, we have developed a Bishnupriya Manipuri Corpus. For this corpus, we have collected the texts from the blogs, websites. We have also collected the texts from the Bishnupriya Manipuri version of Wikipedia. But, there exists a few numbers of resources available on the Internet. Moreover, most of the resources present in the Internet are in image format, we do not have the sophisticated Optical Character Recognition (OCR) tool to extract text format.
Further we have collected large number of text written in Smriti Font, which is an ASCII Font. We have developed an ASCII-to-UNICODE Converter to convert Smriti font texts to UTF-8 texts. Presently, this corpus contains approx. one lakh words, with 10,196 sentences.

## 5  Experiments

We have developed the morphological analyser using XFST tool developed by Xerox. It supports UTF-8 character coding which is important for the implementation of Bishnupriya Manipuri computational morphologies. The tool is based on a lexicon and a set of rules for root and morphemes. This lexicon contains the list of root words and its category separated by a tab. The analyser fails on giving a complex word as an input and the corresponding root word does not exist in the lexicon file. We have developed the Bishnupriya Manipuri lexicon and the rules file required for analysis. The lexicon is designed to reflect the word categories in the Bishnupriya Manipuri language.

The lexicon contains different states for each of the root words, starting with the declaration of the tags.

```
Multichar_Symbols
```

```
                 +Pl +Sg +CM +DM
```

The declaration of tags is followed by the different LEXICONS. For instance, the states for noun can be encoded as-
```
LEXICON Nouns
      মামা N;
LEXICON N
      %+Noun:0 NPL;
      %+Noun:0 PostPositional;
LEXICON NPL
      +Pl:গাছি NCM;
      +Pl:গুলি NCM;
      +Sg:0 NCM;
LEXICON NCM
      +CM:কে NEM;
      +CM:য় NEM;
      +CM:রে NEM;
LEXICON NEM
      +EM:হে #;
      +EM:ক #;
```

The root words and its category are separated by a semicolon. The left side of the colon represents the upper side or the analysis form of the transducer, and the right side shows the lower side or the surface form. The hash symbol at the end of a row indicates the end of the transition, and therefore, that state is the final state. The analyser takes the surface form as input and produces the result as the grammatical structure of the word or the lexicon form.

In the word formation, the morphemes can only appear in certain order and in certain combinations. For example, the initial ই (i) of the suffix ইলু (ilu) is dropped when combine with the roots ending in consonants. Therefore, the following root word কর (kɔɹ : *to do*) changes:

করলু (kɔɹlu) → কর (kɔɹ) + ইলু (ilu)

Again, the ending আ or া (a) in পা (**pa**: *to get*) changes to এ or ে (e) when combine with ইলে (ile).

পেইলে (peile) → পা (pa) + ইলে (ile)

To implement these changes in xfst, we have to include the phonological rules in a separate file with an extension *.regex*. Regular expressions are generally used to manipulate the phonological rules. In the case of পেইলে (**peile**), the phonological rule will be:

[আ | া → এ | ে || _ ই]

which translates as: আ or া becomes এ or ে iff ই follows আ or া.

Figure 1 describes the architecture of used approach.

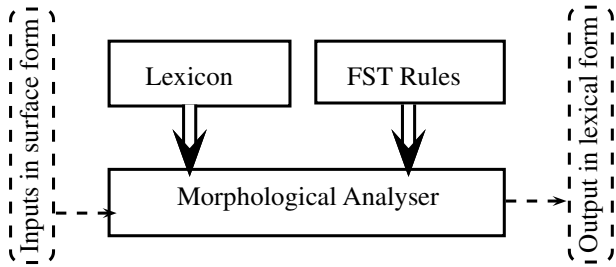

**Fig. 1.** Pictorial work-flow of our approach.

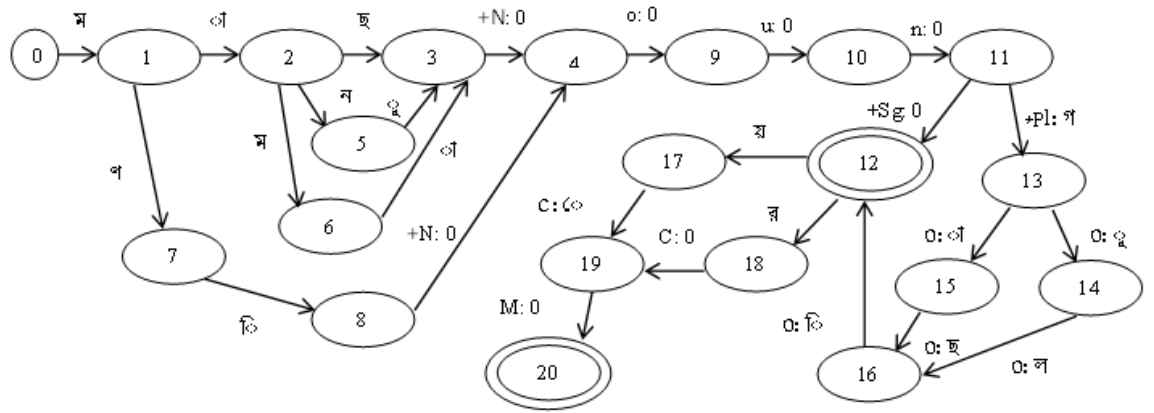

**Fig. 2.** Diagrammatic representation of example words.

## 6 Result & Analysis

We have tested the developed rules with 1000, 2000 and 3000 texts separately. Some of the rules for the noun are given below:

1. Noun + Pl
2. Noun + Pl + CM + EM
3. Noun + PDM
4. Noun + PDM + CM + EM
5. Noun + Sg + SDM + CM + EM

where Pl→Plural Marker, Sg→Singular, CM→Case Marker, EM→Emphatic Marker, PDM→Plural Definitive Marker and SDM→Singular Definitive Marker.

The texts used in this work are part of Bishnupriya Manipuri corpus consisting of web blogs, web sites, Wikipedia and some typed text. We have 150 noun and 50 verb roots as lexicon, in the rule file. After manual verification on the output of the tests, table 3 gives the obtained results with different test data. The experiment was evaluated on the basis of precision and recall values.

Precision: For the morphological analysis, the Precision value is defined as the ratio of the number of words correctly generated to the total number of inputs that should have correctly generated.

Recall: The Recall value is defined as the ratio of the number of words generated to the total number of input words.

F-Score: The F-Score value is define as F= ((2*Precision*Recall)/ (Precision + Recall))

| Test data size | Precision | Recall | F-Score |
|---|---|---|---|
| 1000 | 93.24 | 41.24 | 56.91 |
| 2000 | 91.13 | 38.01 | 53.64 |
| 3000 | 91.03 | 39.38 | 54.98 |

**Table 3.** Obtained results

Figure 2 illustrates the transitions of five words, viz. মামাগাছি (mamagaʃi : *maternal uncles*), মানুয়ে (manuje), মানুর (manuɹ), মাছগুলি (maʃɔguli : *a few fishes*), মণি (mɔni : *jewel*). The output of these five words is as below:

মামাগাছি→মামা+Noun+Pl
মানুয়ে→মানু+Noun+Sg+CM
মানুর→মানু+Noun+Sg+CM
মাছগুলি→মাছ+Noun+Pl
মণি→মণি+Noun+Sg

The automation for these five words is represented as a finite state transducer: a finite set of states together with a set of arcs between pairs of states. States are represented as circles and arcs are represented by arrows from one state to another state. The final states are represented by two concentric circles. The morphological analyser starts at the initial state and goes through a sequence of states by computing the morphemes. If it matches the symbol on an arc leaving the present state, then it moves to the next state through that arc and moves to the next symbol of the input word. Thus it successfully recognizes all the morphemes in an input string.

# 7 Conclusion

This paper has described the morphological analyser for the Bishnupriya Manipuri language based on the finite state transducer. The designed rules are successfully tested with the text collected from the Bishnupriya Manipuri Corpus. Our main motivation is to develop the resources of linguistics work on this language and being the first work of this kind for the Bishnupriya Manipuri language, we obtained a good precision for our work.